\pdfoutput=1

\documentclass[11pt]{article}

\usepackage[final]{acl}

\usepackage{times}
\usepackage{latexsym}

\usepackage[T1]{fontenc}

\usepackage[utf8]{inputenc}

\usepackage{microtype}

\usepackage{inconsolata}

\usepackage{graphicx}
\usepackage{comment}

\usepackage{array}
\usepackage{multirow}
\usepackage{booktabs}
\usepackage{amsmath}    
\usepackage{amssymb}    
\usepackage{bm}         
\usepackage[most]{tcolorbox}
\usepackage{todonotes}

\usepackage{booktabs}
\usepackage{graphicx}
\usepackage{multirow}
\usepackage{subcaption}

\title{Zero-Shot Cross-Lingual Transfer using Prefix-Based Adaptation}

\author{
 \textbf{Snegha A\textsuperscript{$\ddagger$}},
 \textbf{Sayambhu Sen\textsuperscript{$\S$}},
  \textbf{Piyush Singh Pasi\textsuperscript{$\S$}},
 \textbf{Abhishek Singhania\textsuperscript{$\S$}},
 \\
 \textbf{Preethi Jyothi\textsuperscript{$\ddagger$}}
\\
\\
 \textsuperscript{$\ddagger$}
Indian Institute of Technology Bombay, India,\\
 \textsuperscript{$\S$}Amazon Alexa 
 \\
\texttt{\normalsize{\{23m2160,pjyothi\}@iitb.ac.in, \{sensayam,piyushpz,mrabhsin\}@amazon.com}
 }
}

\begin{document}
\maketitle
\begin{abstract}
With the release of new large language models (LLMs) like Llama and Mistral, zero-shot cross-lingual transfer has become increasingly feasible due to their multilingual pretraining and strong generalization capabilities. However, adapting these decoder-only LLMs to new tasks across languages remains challenging. While parameter-efficient fine-tuning (PeFT) techniques like Low-Rank Adaptation (LoRA) are widely used, prefix-based techniques such as soft prompt tuning, prefix tuning, and Llama Adapter are less explored, especially for zero-shot transfer in decoder-only models. We present a comprehensive study of three prefix-based methods for zero-shot cross-lingual transfer from English to 35+ high- and low-resource languages. Our analysis further explores transfer across linguistic families and scripts, as well as the impact of scaling model sizes from 1B to 24B. With Llama 3.1 8B, prefix methods outperform LoRA-baselines by up to \textbf{6\%} on the Belebele benchmark. Similar improvements were observed with Mistral v0.3 7B as well. Despite using only 1.23M learning parameters with prefix tuning, we achieve consistent improvements across diverse benchmarks. These findings highlight the potential of prefix-based techniques as an effective and scalable alternative to LoRA, particularly in low-resource multilingual settings.

\end{abstract}
\section{Introduction}
Large language models (LLMs) exhibit strong multilingual and zero-shot generalization abilities due to exposure to diverse pretraining data. Nonetheless, cross-lingual transfer remains challenging given the linguistic diversity and complexity of adapting large models efficiently without significant computational overhead.

To address the high computational and memory costs of full model finetuning, recent advances in parameter-efficient finetuning (PeFT) techniques focus on updating only a small subset of model parameters while keeping the majority of the pretrained weights frozen. This design significantly reduces the adaptation cost and makes large-scale models more practical for multilingual and domain-specific applications. Methods such as Low-Rank Adaptation (LoRA)~\cite{hu2022lora} and instruction-tuned adapters have shown promising results in efficiently tailoring models to new tasks without requiring extensive resources. Among the various PeFT techniques, prefix-based approaches like soft prompting~\cite{lester-etal-2021-power, liu2024gpt} and prefix tuning~\cite{li-liang-2021-prefix} are particularly compelling because they introduce learnable components either at the input or within the transformer stack, enabling flexible task adaptation without altering the underlying architecture of the model.  

While these prefix-based techniques have been shown to be effective in monolingual scenarios and task-specific settings, their potential in facilitating zero-shot cross-lingual transfer is under-explored. This is especially relevant for decoder-only LLMs, which are increasingly being deployed in multilingual environments. Unlike encoder-decoder models that have been more thoroughly studied for transfer across languages, decoder-only models present unique challenges due to their reliance on autoregressive decoding. Understanding whether prefix-based PeFT methods can enhance zero-shot cross-lingual performance in such models has not been previously studied in detail. 

In this work, we provide the first systematic study of prefix-based PeFT methods for zero-shot cross-lingual transfer in \emph{decoder-only LLMs}. Our contributions can be summarized as follows:

\begin{itemize}
    \item We evaluate prefix-based adaptation on models ranging from 1B parameters to large-scale 24B models to show the effectiveness of prefix tuning in multilingual transfer across models of varying sizes.  
    
    \item Our study spans four well-recognized multilingual benchmarks -- XQUAD, XNLI,  Belebele and MGSM -- to compare the performance of LoRA and prefix-based tuning.  
    
    \item We provide a detailed comparison of prefix-based methods (soft prompts, prefix tuning, LLaMA-Adapter) against LoRA and full fine-tuning\footnote{Due to computational limitations, full fine-tuning is restricted to the SQuAD dataset on Llama 3.1 8B.}, systematically analyzing their strengths and limitations across tasks and 35+ high- and low-resource languages. Additionally, we investigate transfer patterns across linguistic families and scripts. 
\end{itemize}

Together, our findings position prefix-based adaptation as a lightweight yet powerful strategy for cross-lingual and reasoning-oriented applications, particularly in resource-constrained multilingual settings.

\begin{figure*}
    \centering
    \includegraphics[width=1\linewidth]{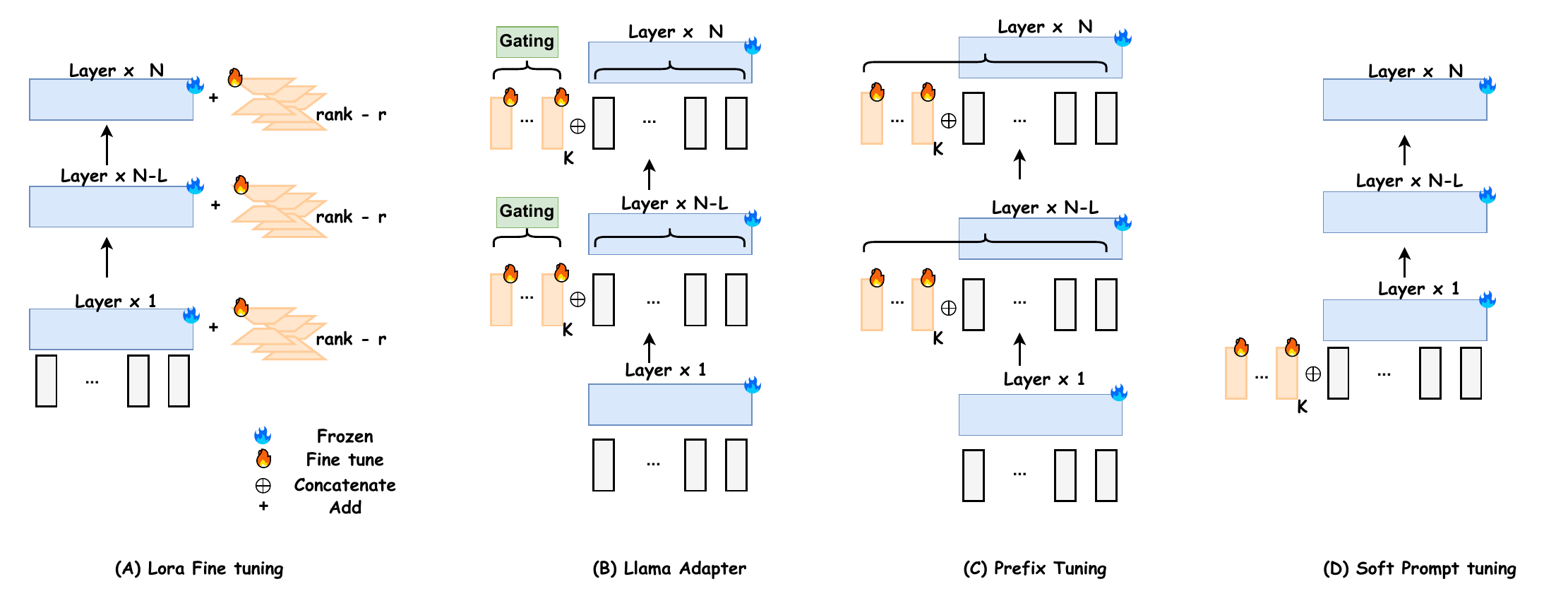}
    \caption{Schematic representation of: (A) LoRA fine-tuning and prefix-based methods, (B) Llama Adapter, (C) Prefix tuning, and (D) Soft prompt tuning.}
    \label{fig:enter-label}
\end{figure*}

\section{Related Work} 

Cross-lingual transfer is a key challenge in multilingual NLP. It is traditionally tackled through full fine-tuning of multilingual models. However, with large decoder-only LLMs like Llama and Mistral, full fine-tuning is costly, leading to PeFT approaches. LoRA \cite{hu2022lora} introduces low-rank trainable matrices into frozen weights to reduce training overhead. Alternatively, prefix-based methods either add learnable tokens at the input layer \cite{lester-etal-2021-power, liu2024gpt} or to attention keys and values at each layer \cite{li-liang-2021-prefix}, enabling efficient task adaptation.

Soft prompt tuning has been extensively studied for cross-lingual transfer in encoder and encoder-decoder models, particularly in classification tasks. For instance, \cite{philippy-etal-2024-soft} demonstrated that soft prompts can generalize better across languages with fewer parameters, following the ``less is more" principle. Similarly, \cite{philippy-etal-2025-enhancing} utilized multilingual verbalizers and contrastive label smoothing to further enhance cross-lingual classification. Recent work such as \cite{vykopal-etal-2025-soft} introduced language-specific soft prompts specifically designed for transfer learning, showing that combining language-specific and task-specific prompts improves generalization. However, these prior works  predominantly used multilingual encoder-only and encoder-decoder models, and appended prefix tokens only to the input. 

As soft prompts have several limitations in effectively adapting models to new tasks, prefix tuning emerged as a promising approach. Cross-lingual alignment through prompt-based pretraining, as proposed by \cite{tu-etal-2024-efficiently}, further improved intent classification and slot-filling performance but it is not a zero-shot setting (as in our work).  A recent variant of prefix tuning is LLaMA Adapter \cite{dubey2024llama} that introduced zero-initialized attention mechanisms for efficient prefix training and achieved strong instruction-following capabilities; however, they did not evaluate on any multilingual benchmarks. A related line of work has focused on extending prefix tuning to instance-specific adaptation based on the input prompt for improved model performance~\cite{liu-etal-2024-para, jiang2022instance, liu2024intuition, zhu-etal-2024-iapt}. 

Few comparative studies have examined parameter-efficient tuning for multilingual settings, and most have been restricted to encoder-only models or small decoder-only models with only a few million parameters. For instance, \cite{zhao-schutze-2021-discrete} systematically compared discrete prompting, soft prompting, and fine-tuning on the few-shot multilingual NLI task using XLM-RoBERTa-base. Similarly, \cite{tu-etal-2022-prompt} compared prompt tuning with fine-tuning across diverse NLU tasks on XLM-R and mBERT. \cite{tu-etal-2022-prompt} evaluate prefix tuning on the encoder-only XLM-R model and showed its effectiveness over full fine-tuning in zero-shot cross-lingual transfer. ~\citet{tu-etal-2022-prompt} investigated a decoder-based multilingual model (XGLM), but their analysis was limited to a single small model. They showed that prompt tuning can sometimes surpass fine-tuning, particularly for low-resource languages, although performance remained highly sensitive to the underlying tokenization scheme. Our work significantly extends their analysis to large decoder-only LLMs and presents a comprehensive comparison of multiple prefix-based methods, including soft prompts, prefix tuning, and LLaMA Adapter. 

\section{Methodology}
\paragraph{Low-Rank Adaptation (LoRA).}  
LoRA \cite{hu2022lora} is a parametric fine-tuning technique that has become one of the most popular approaches to enable cross-lingual transfer in LLMs. It introduces trainable low-rank matrices, typically in the query, key, and value projections, while keeping the base model frozen. These learned matrices are added to the original weights during inference. Unlike prefix-based methods, LoRA directly modifies the model parameters. A standard cross-lingual transfer setup involves fine-tuning the model using LoRA on task-specific English data and evaluating it on the target language of interest. Formally, let \( W \in \mathbb{R}^{d \times k} \) be a pretrained weight matrix of a projection layer (e.g., \( W_q, W_k, W_v \)). Instead of updating \( W \) directly, LoRA parameterizes the weight update as a product of two low-rank matrices:  
\begin{equation}
\Delta W = B A, \quad A \in \mathbb{R}^{r \times k}, \, B \in \mathbb{R}^{d \times r}, \label{eq:lora_update}
\end{equation}

\noindent where \( r \ll \min(d, k) \) is the rank of the adaptation. The modified projection becomes:  

\begin{equation}
W' = W + \Delta W = W + B A. \label{eq:lora_weight}
\end{equation}

Given an input hidden state \( h \in \mathbb{R}^{k} \), the output of the adapted projection layer is computed as:  

\begin{equation}
y = W' h = W h + BA h. \label{eq:lora_forward}
\end{equation}

Here, only \( A \) and \( B \) are trainable, while \( W \) remains frozen. This formulation enables efficient fine-tuning by reducing the number of trainable parameters and allowing task-specific adaptation without updating the full weight matrices.

\begin{table*}[t]
\centering
\small
\resizebox{\textwidth}{!}{
\begin{tabular}{l|ccccccccccccccc|c}
\toprule
\textbf{Method} & \textbf{en} & \textbf{hi} & \textbf{el} & \textbf{vi} & \textbf{sw} & \textbf{bg} & \textbf{th} & \textbf{ar} & \textbf{de} & \textbf{es} & \textbf{fr} & \textbf{ru} & \textbf{tr} & \textbf{zh} & \textbf{ur} & \textbf{Avg}\\
\midrule
Base Model       & 53.8 & 48.0 & 51.0 & 49.7 & 45.9 & 52.0 & 48.0 & 48.6 & 50.4 & 51.8 & 49.6 & 50.8 & 50.0 & 50.1 & 48.7 & 49.9\\
LoRA$_4$  & 90.3 & 70.1 & 74.5 & 77.5 & \underline{60.2} & 73.0 & 72.4 & 71.5 & 77.8 & 79.2 & 80.2 & 73.6 & 73.2 & 77.9 & 65.0 & 74.4\\
Soft Prompts   & 84.3 & 67.9 & 54.2 & 72.7 & 51.4 & 51.4 & 66.1 & 63.2 & 52.3 & 57.2 & 59.0 & 59.4 & 58.4 & 41.4 & 62.6 & 60.1\\
Llama Adapter  & \underline{93.4} & \underline{74.5} & \textbf{79.8} & \textbf{79.2} & 59.6 & \underline{78.4} & \underline{76.0} & \textbf{76.5} & \textbf{83.8} & \underline{83.7} & \underline{84.6} & \underline{79.6} & \textbf{75.9} & \textbf{81.2} & \textbf{71.8} & \underline{78.5}\\
Prefix Tuning  & \textbf{93.9} & \textbf{76.5} & \underline{79.4} & \underline{79.1} & \textbf{60.3} & \textbf{79.4} & \textbf{76.2} & \underline{75.7} & \underline{83.5} & \textbf{84.4} & \textbf{85.0} & \textbf{79.9} & \underline{75.5} & \underline{79.9} & \underline{71.7} & \textbf{78.7}\\
\bottomrule
\end{tabular}
}

\caption{LLaMA 3.1 8B performance (accuracy) on XNLI benchmark comparing LoRA and prefix based adaption methods.The best performance for each language is shown in \textbf{bold}, and the second-best is \underline{underlined}.}
\label{tab:task-b}
\end{table*}

\begin{table*}[t]
\centering
\small
\begin{tabular}{l|cccccccccccc|c}
\toprule
\textbf{Method} & \textbf{en} & \textbf{hi} & \textbf{el} & \textbf{vi} & \textbf{ar} & \textbf{de} & \textbf{es} & \textbf{ro} & \textbf{ru} & \textbf{th} & \textbf{tr} & \textbf{zh} & \textbf{Avg} \\
\midrule
Base Model       & 79.3 & 59.3 & 60.5 & 71.2 & 59.4 & 68.5 & 67.6 & 68.5 & 60.3 & 63.2 & 62.5 & 59.5 & 65.0\\
LoRA$_4$    & 86.2 & 66.1 & 72.0 & 75.3 &  68.9 & 76.4 & 78.0 & 78.0 & 72.0 & \textbf{75.8} & 69.1 & 71.7 & 74.1\\
Soft Prompts   & 54.5 & 10.8 & 27.9 & 45.5 & 25.8 & 42.2 & 52.0 & 48.2 & 32.2 & 11.2 & 36.8 & 16.1 & 33.6\\
Llama Adapter  & \underline{89.4} & \underline{75.1} & \underline{76.9} & \underline{79.8} & \textbf{72.1} & \underline{82.4} & \underline{83.2} & \underline{82.6} & \textbf{78.4} & \underline{71.6} & \textbf{73.3} & \underline{72.3} & \textbf{78.1}\\
Prefix Tuning & \textbf{90.2} & \textbf{75.7} & \textbf{78.4} & \textbf{79.3} & \underline{70.4} & \textbf{82.8}& \textbf{84.2} & \textbf{83.5} & \underline{76.9} & 70.9 & \underline{72.6} & \textbf{72.5} & \textbf{78.1}\\

\bottomrule
\end{tabular}
\caption{Llama 3.1 8B performance (F1 score) on XQUAD benchmark comparing LoRA and prefix based adaption methods. The best performance for each language is shown in \textbf{bold}, and the second-best is \underline{underlined}.}
\label{tab:task-c2}
\end{table*}

\begin{table*}[t]
\centering
\small
\begin{tabular}{l|ccccccccccc|c}
\toprule
\textbf{Method} & \textbf{en} & \textbf{th} & \textbf{zh} & \textbf{sw} & \textbf{fr} & \textbf{bn} & \textbf{de} & \textbf{te} & \textbf{ja} & \textbf{es} & \textbf{ru} & \textbf{Avg} \\
\midrule
Base Model       & 50.4 & 23.2 & \underline{27.6} & \textbf{13.2} & 28 & \textbf{16.4} & 26 & \textbf{12.4} & 16.8 & 34.8 & 30 & 25.34\\
LoRA$_4$    & 36.8 & 16.8 & 27.6 & 7.6 &  25.2 & 4.8 & 22.8 & 0.8 & 19.2 & 24 & 27.2 & 19.34\\
Llama Adapter  & \textbf{53.6} & 18.4 & 32.4 & 8 & \underline{32.8} & 9.6 & \underline{33.6} & 2 & \underline{25.2} & \underline{35.6} &  \underline{32} & \underline{25.74}\\
Prefix Tuning & \underline{52.8} & \textbf{26} & \textbf{37.6} & \underline{10.8} & \textbf{34} & \underline{12.8}& \textbf{41.2} &  \underline{6.4} & \textbf{25.6} & \textbf{37.6} & \textbf{39.2} & \textbf{29.45}\\

\bottomrule
\end{tabular}
\caption{Llama 3.1 8B performance (maj@1) on MGSM benchmark comparing LoRA and prefix based adaption methods. The best performance for each language is shown in \textbf{bold}, and the second-best is \underline{underlined}.}
\label{tab:mgsm}
\end{table*}

\textbf{}

\paragraph{Prefix Tuning.}  

Given an LLM, prefix tuning \cite{li-liang-2021-prefix} introduces a set of learnable prefix tokens to all layers of the transformer. In our implementation, we only append the learnable prefixes to the final \( L \) layers of the transformer. The main intuition is that these prefix tokens act as additional context vectors that the model can attend to. These vectors guide the model toward task-specific behavior, while the pretrained parameters of the LLM remain frozen. 

Formally, let \( P_l \in \mathbb{R}^{K \times d} \) denote the learnable prefix tokens at layer \( l \), where \( K \) is the number of prefix tokens and \( d \) is the embedding dimension. We consider the computation for the \( (M+1) \)-th token, denoted by \( t_l \in \mathbb{R}^{1 \times d} \).The layer’s input hidden states (including the current token) are represented as \( H_l \in \mathbb{R}^{(M+1) \times d} \). Each attention head operates on these hidden states using projection matrices \( W_q, W_k, W_v \in \mathbb{R}^{d \times d} \).

The query vector corresponding to the current token \( t_l \) is computed using the frozen projection matrix \( W_q \):
\begin{equation}
Q_{l} = t_{l} \, W_{q} \in \mathbb{R}^{1 \times d} \nonumber
\end{equation}

The keys and values corresponding to the input sequence (\( H_l \)) are also computed using the frozen projection matrices \( W_k \) and \( W_v \):
\begin{equation}
K^{H}_{l} = H_{l} \, W_{k}, \quad V^{H}_{l} = H_{l} \, W_{v} \nonumber
\end{equation}

The key idea of prefix tuning is the concatenation of the learnable prefix parameters with keys and values derived from the input. 
\begin{equation}
P^{K}_{l} = P_{l} \, W_{k}, \quad P^{V}_{l} = P_{l} \, W_{v} \nonumber
\end{equation}
\( P^{K}_{l} \in \mathbb{R}^{K \times d} \) and \( P^{V}_{l} \in \mathbb{R}^{K \times d} \) denote the learnable prefix keys and values of layer \( l \), respectively. The final keys and values at layer l become:

\begin{equation}
K_{l} = [P^{K}_{l}; K^{H}_{l}], \quad V_{l} = [P^{V}_{l}; V^{H}_{l}] \nonumber
\end{equation}
\( K_l \) and \( V_l \) are expanded matrices encompassing both the learned prefixes and the input sequence.

The attention scores are obtained by comparing the query \( Q_l \) against the concatenated keys \( K_l \):
\begin{equation}
S_{l} = \frac{Q_{l} K_{l}^{T}}{\sqrt{d}} \in \mathbb{R}^{1 \times (K + M + 1)}. \label{eq:attention_scores}
\end{equation}

\noindent The attention distribution is computed by applying the softmax function, which weights the contributions of both the prefix and the input tokens:
\begin{equation}
A_{l} = \text{softmax}(S_{l}) = \left[ A^{P}_{l}, A^{H}_{l} \right] \in \mathbb{R}^{1 \times (K + M + 1)}, \nonumber
\end{equation}
where \( A^{P}_{l} \) represents the attention weights over the learned prefixes and \( A^{H}_{l} \) represents the weights over the input sequence.

Finally, as is typically done in transformer models, the attended output representation at layer \( l \) is computed as a weighted sum of the concatenated values \( V_l \), followed by an output projection:
\begin{equation}
t^{o}_{l} = (A_{l} V_{l}) \, W_{o} \in \mathbb{R}^{1 \times d}, \label{eq:output}
\end{equation}
where \( W_o \) is the output projection matrix. In this way, prefix tuning directly modifies the attention mechanism by injecting learned keys and values (\( P^{K}_{l}, P^{V}_{l} \)), steering the model’s representations without modifying the base model weights.

\paragraph{Llama Adapter.}  
The Llama Adapter \cite{zhang2024llamaadapter} builds upon the principles of prefix tuning but introduces an important modification to stabilize training in large-scale LLMs. Specifically, it replaces the standard attention mechanism with a zero-initialized variant. This mitigates instabilities that often arise from randomly initialized prefix tokens in the early stages of fine-tuning. To further enhance stability, a learnable gating mechanism is introduced, allowing the model to gradually scale the influence of prefix tokens during optimization. The gated attention scores are given by:
\begin{equation}
A^{g}_{l}=\big[\text{softmax}(S^{K}_{l})\cdot\tanh(g_{l}),\,\text{softmax}(S^{M+1}_{l})\big] \label{eq:gated_scores}
\end{equation}

\noindent where the attention scores can be split into contributions from the learnable prefix \( S^{K}_{l} \) and the original sequence \(S^{M+1}_{l}\). \( g_l \) is a learnable scalar gating that adaptively controls the contribution of the prefix tokens. Finally, the output representation \( t^{o}_{l} \) is obtained using the same formulation in Equation~\ref{eq:output}. By weighting the prefix contributions using a learned gate,  Llama Adapter ensures stable and effective adaptation of decoder-only LLMs.  

\paragraph{Soft Prompts.} Soft prompts~\cite{lester-etal-2021-power, liu2024gpt} involve prepending learnable continuous embeddings to the input, serving a similar goal as manual prompts. However, instead of manually selecting discrete prompts, soft prompting optimizes a continuous set of embeddings that serve as the prompt. This allows the model to learn how to best steer its behavior through gradient-based updates to the soft prompts. 

Let $S \in \mathbb{R}^{K \times d}$ represent the learnable 
soft prompt embeddings, where $K$ denotes the number of prompt tokens and 
$d$ is the hidden dimension. Given an input sequence $T$, the modified input 
$\tilde{T}$ is obtained by prepending the soft prompts:
\begin{equation}
    \tilde{T} = [S; T]
\end{equation}
where $[;]$ denotes concatenation. The sequence $\tilde{T}$ is then passed 
through the transformer as usual, with $S$ updated via gradient-based optimization 
during fine-tuning.
Unlike prefix tuning, which injects key-value pairs at every transformer layer, soft prompting only modifies the input embeddings.

\section{Experiments}
\paragraph{Models.} All experiments are conducted on Llama 3.1 (8B) \cite{dubey2024llama} and Mistral v0.3 (7B) \cite{jiang2023mistral}.To study the effect of model scaling, we additionally evaluate smaller and larger variants - Llama 3.2 (1B) and Mistral Small (24B), respectively.
 The Llama 3.1 and 3.2 series, developed by Meta, comprise multilingual large language models. Mistral v0.3 (7B) is an updated release from Mistral AI with an extended vocabulary compared to Mistral v0.1. Notably, Mistral Small (24B) establishes a new benchmark in the ``small" LLM category (under 70B) by offering improved multilingual capabilities and a larger vocabulary. We have limited our experiment to the base model variants only.

\paragraph{Datasets.} We evaluate on three widely-used cross-lingual benchmarks, each targeting a distinct aspect of language understanding: XQUAD \cite{Artetxe:etal:2019} for cross-lingual question answering, \textsc{XNLI} \cite{conneau2018xnli} for cross-lingual natural language inference, and Belebele \cite{bandarkar-etal-2024-belebele} for cross-lingual machine reading comprehension. We also evaluate on the MGSM \cite{shi2023language} benchmark to assess the reasoning capabilities of large language models in multilingual settings.

\paragraph{Training Details}
\label{sec:training_parameter}
We fine-tune prefix-based adaptation methods and LoRA with rank 4 using the English SQuAD training set for XQUAD containing 87.6K samples and a subset of the English NLI training data containing 100K samples for XNLI evaluations. For Belebele, we use their suggested training set containing 67.5K English samples. Finally, we use the GSM8K English training dataset with 7.47K samples \cite{cobbe2021training} and evaluate on MGSM. All the datasets are publicly available; more training details are in Appendix~\ref{sec:prompt_template}. 

We experimented with learning rates (3e-3, 1e-3 and 3e-4), epochs (2,3,5), and weight decay (0.02,0.04,0.1), and report the best performance for each model. We used a learning rate of 3e-3, 2 epochs, and a weight decay of 0.02. For XNLI, we sampled 1,000 instances per language for evaluation due to computational constraints. Since XQUAD does not provide a separate test set, we evaluated on the full validation set, which includes approximately 1.19K samples per language. Finally for Belebele, we evaluated on 23 languages, where each language has 900 samples. All experiments were conducted on a single NVIDIA A100 80GB GPU.

\begin{table*}[t]
\centering
\caption{Overall Llama 3.1 8B performance (accuracy) on the Belebele benchmark, grouped by script and family. Best performance is in \textbf{bold}, second-best is \underline{underlined}.}
\label{tab:main_performance_table}
\begin{minipage}{0.49\textwidth}
\centering
\resizebox{\textwidth}{!}{
\begin{tabular}{l|l|ccccc}
\toprule
\textbf{Script} & \textbf{Language} & \textbf{Base} & \textbf{LoRA$_4$} & \textbf{Soft} & \textbf{Llama} & \textbf{Prefix} \\
\textbf{} & \textbf{} & \textbf{Model} & \textbf{} & \textbf{Prompt} & \textbf{Adapter} & \textbf{tuning} \\
\midrule
\multirow{3}{*}{Cyrillic} 
 & Kyrgyz    & 37.2 & 52.9 & 59.3 & \underline{60.5} & \textbf{64.2} \\
 & Russian   & 50.4 & 81.0 & 86.1 & \underline{87.7} & \textbf{88.1} \\
 & Serbian   & 48.7 & 71.7 & 81.1 & \textbf{81.9} & \underline{81.5} \\
\midrule
\multirow{3}{*}{Burmese} 
 & Burmese    & 30.9 & 36.2 & 43.3 & \underline{45.1} & \textbf{48.4} \\
 & Shan   & 31.1 & 28.0 & 30.0 & 29.0 & \textbf{33.0} \\
\midrule
\multirow{3}{*}{Latin} 
 & Swati    & 30.2 & \underline{34.3} & 33.4 & \underline{34.3} & \textbf{34.5} \\
 & Sundanese   & 35.3 & 47.1 & 52.3 & \underline{56.4} & \textbf{57.8} \\
 & Bambara   & 28.4 & \textbf{34.3} & \underline{33.1}  & 32.2 & 32.2 \\
\midrule
\multirow{3}{*}{Arabic} 
 & Sindhi    & 36.9 & 46.4 & 51.1 & \underline{53.3} & \textbf{55.8} \\
 & Egyptian Arabic   & 40.1 & 57.6 & 65.2 & \underline{68.4} & \textbf{68.7} \\
 & Western Persian   & 47.5 & 72.9 & 79.6 & \underline{81.4} & \textbf{82.2} \\
\midrule
\multirow{3}{*}{Ethiopic}
 & Amharic     & 30.5 & 34.7 & \underline{37} & 34.9 & \textbf{37.8} \\
 & Tigrinya    & 24 & 29.2 & \underline{29.7} & 28.1 & \textbf{29.8} \\
\bottomrule
\end{tabular}
}
\subcaption{Grouped by language \textbf{script}.}
\label{tab:grouped_by_script}
\end{minipage}
\hfill
\begin{minipage}{0.49\textwidth}
\centering
\resizebox{\textwidth}{!}{
\begin{tabular}{l|l|ccccc}
\toprule
\textbf{Family} & \textbf{Language} & \textbf{Base} & \textbf{LoRA$_4$} & \textbf{Soft} & \textbf{Llama} & \textbf{Prefix} \\
\textbf{} & \textbf{} & \textbf{Model} & \textbf{} & \textbf{Prompting} & \textbf{Adapter} & \textbf{tuning} \\
\midrule
\multirow{3}{*}{Turkic}
 & Kazakh            & 38 & 53.8 & 61.8 & \underline{63.9} & \textbf{64.2} \\
 & Kyrgyz            & 37.2 & 52.9 & 59.3 & \underline{60.5} & \textbf{64.2} \\
 & North Azerbaijani & 39.9 & 58.4 & 65.4 & \underline{68.3} & \textbf{68.5} \\
\midrule
\multirow{3}{*}{Dravidian}
 & Kannada           & 35.2 & 46.0 & 59.3 & \underline{59.5} & \textbf{61.1} \\
 & Malayalam         & 35.5 & 49.3 & 56.9 & \underline{60.0} & \textbf{63.9} \\
 & Tamil             & 36.9 & 52.3 & 60.1 & \underline{60.8} & \textbf{65.3} \\
\midrule
\multirow{3}{*}{Afro-Asiatic}
 & Amharic            & 30.5 & 34.7 & \underline{37.0} & 34.9 & \textbf{37.8} \\
 & Tigrinya            & 24 & 29.2 & \underline{29.7} & 28.1 & \textbf{29.8} \\
 & Tsonga      & 32.7 & 36.3 & \underline{37.3} & 36.1 & \textbf{39} \\
\midrule
\multirow{3}{*}{Indo-Aryan}
 & Sindhi            & 36.9 & 46.4 & 51.1 & \underline{53.3} & \textbf{55.9} \\
 & Odia            & 33.1 & 38.2 & 54.7 & \underline{55.3} & \textbf{59.1} \\
 & Sinhala & 34.2 & 47.8 & \underline{54.8} & 53.8 & \textbf{60.8} \\
 \midrule
\multirow{3}{*}{Balto-Slavic}
 & Russian   & 50.4 & 81.0 & 86.1 & \underline{87.7} & \textbf{88.1} \\
& Serbian   & 48.7 & 71.7 & 81.1 & \textbf{81.9} & \underline{81.5} \\
 & Slovak    & 46.5 & 73.8 & 80.6 & \underline{83.5} & \textbf{84.3 }\\
\bottomrule
\end{tabular}
}
\subcaption{Grouped by language \textbf{family}.}
\label{tab:grouped_by_family}
\end{minipage}
\end{table*}

\section{Analysis and Ablations}

\paragraph{Comparison with LoRA Fine-Tuning.}  Tables ~\ref{tab:task-b} and ~\ref{tab:task-c2} (and Tables ~\ref{tab:task-a} and ~\ref{tab:task-c1} in the Appendix) shows the performance of Llama 3.1 and Mistral v0.3 models across various tuning strategies, including LoRA, soft prompt tuning, prefix tuning, and Llama adapters on the \textsc{XNLI} and XQUAD datasets. To ensure fair comparisons, the number of trainable parameters in LoRA was matched with those of the prompt-based methods by setting $r=4$ and $\alpha=8$. The results show that prefix-based methods consistently outperform LoRA on both LLama 3.1 8B and Mistral v0.3 7B with English as the source language. This highlights the ability of prefix-based tuning for effective multilingual adaptation, even with as little as \textbf{1.23M} model parameters being trained.


We observe consistent improvements from prefix tuning across all benchmarks. Using Llama 3.1 (8B), prefix tuning achieves up to \textbf{28\%} higher accuracy on XNLI, \textbf{13\%} higher F1 on XQUAD, and \textbf{18\%} higher accuracy on Belebele compared to the base model. Moreover, it provides additional gains of up to \textbf{4--6\%} over LoRA, as shown in Tables~\ref{tab:task-b}, \ref{tab:task-c2}, \ref{tab:grouped_by_script}, and~\ref{tab:grouped_by_family}. Similar trends are observed for Mistral, with consistent improvements reported in Tables, ~\ref{tab:family_mistral} and~\ref{tab:script_mistral} in Appendix~\ref{sec:Final_numbers}.


\paragraph{Effectiveness of prefix-based methods across high and low-resource languages.} We further evaluate the effectiveness of prefix-based methods on languages categorised as high and low resource. Since \textsc{XNLI} and XQUAD benchmarks primarily span high-resource languages, we rely on the Belebele benchmark to assess performance on low-resource languages. We select 23 languages for our analysis, of which 19 are considered low-resource and 4 high-resource, as per the \textsc{FLoRes} dataset classification. Across both the Mistral and Llama architectures, prefix-based adaptation methods yield significant performance gains while requiring only \textbf{1.23M} parameters to be tuned. Among low-resource languages, absolute improvements range from a minimum of \textbf{2\%} for Shan to a maximum of \textbf{37\%} for Western Persian using Llama 3.1 8B. 

Prefix tuning and LLaMA adapters typically yield better cross-lingual transfer than soft prompts, likely due to more tunable parameters. However, in low-resource scenarios like those in the Belebele benchmark, soft prompting performs comparably or better as shown in Tables~\ref{tab:main_performance_table} and Tables \ref{tab:family_mistral}, \ref{tab:script_mistral} in Appendix \ref{sec:Final_numbers}. This is likely due to their lightweight design that helps preserve pretrained multilingual knowledge. Overall, prefix based methods appear to leverage inherent language knowledge better than LoRA.

\paragraph{Influence of script and language family.}
From Tables~\ref{tab:grouped_by_script} and \ref{tab:grouped_by_family}, we observe that while both script-wise and family-wise groupings reveal performance gains with prefix-based methods, language family appears to be a reliable indicator of adaptation success. Languages within the same family tend to benefit similarly. Script-based trends show more variability, likely influenced by resource availability and linguistic diversity within a script group. The languages in our analysis span a diverse range of families such as Turkic, Dravidian, Afro-Asiatic, Balto-Slavic, and Indo-Aryan. The scripts span Cyrillic, Burmese, Arabic, Ethiopic, and Latin. Many of these languages are typologically and morphologically distant from our source language English. Prefix-based methods show strong cross-lingual performance even across distant languages, suggesting that typological similarity to English is not essential for effective adaptation. Similar trends are observed with Mistral as well, as shown in Table \ref{tab:family_mistral} and \ref{tab:script_mistral} in Appendix \ref{sec:Final_numbers}.
\captionsetup{font=small}
\begin{table}[t]
\centering
\small
\setlength{\tabcolsep}{4pt}
\begin{tabular}{c|c|c}
\toprule
\textbf{Method} & \textbf{Params} & \textbf{Acc.} \\
\midrule
Full Fine-tuning      & $\sim8B$ & 37.74 \\
LoRA$_4$     & 75.50M & 75.99 \\
Llama Adapter     & 1.23M & 78.09 \\
Prefix tuning     & 1.23M & \textbf{78.11} \\
\bottomrule
\end{tabular}
\vspace{0.5em}
\caption{Comparison of full fine-tuning and parameter-efficient methods on the XQUAD dataset using LLama 3.1 8B, reported in terms of average F1 score across all languages.}
\label{tab:fullft}
\end{table}
\paragraph{Prefix-based adaptation vs. full fine-tuning.}
Table~\ref{tab:fullft} presents a comparison of LoRA, prefix-based methods, and full fine-tuning. Detailed language-wise results are provided in Table \ref{tab:fft_3} in Appendix~\ref{sec:Final_numbers}. We observe that while full fine-tuning leads to improvements in English, it negatively impacts the performance of target languages when applied to decoder-only models such as Llama-3.1 8B. Due to computational constraints, we were unable to extensively tune hyperparameters to achieve the best possible results. Overall, our findings indicate that LoRA and prefix-based methods are more effective and efficient choices for zero-shot cross-lingual transfer compared to full fine-tuning. We hypothesize that this could primarily be due to full-finetuning (on English data) leading to catastrophic forgetting in other languages.


\begin{table}[t]
\small
\setlength{\tabcolsep}{4pt} 
\begin{tabular}{l|ccccccccccccccc|c}
\toprule
\textbf{Method} & \textbf{LLaMA} & \textbf{Mistral} & \textbf{LLaMA} & \textbf{Mistral} \\
 & \textbf{3.2 1B} & \textbf{v0.3 7B} & \textbf{3.1 8B} & \textbf{24B} \\
\midrule
Base Model       & 27.51 & 56.1 & 65.0 & 72.57 \\
LoRA$_4$  & 56.80 & 59.12 & 74.1 & 70.19 \\
Llama Adapter  & \underline{64.26} & \underline{65.1} & \textbf{78.1} & \underline{79.70} \\
Prefix Tuning  & \textbf{64.46} & \textbf{67.2} & \textbf{78.1} & \textbf{79.94} \\
\bottomrule
\end{tabular}
\caption{Average Performance across all languages on XQUAD (F1 score) benchmark across all models comparing LoRA and prefix based adaption methods.The best performance for each language is shown in \textbf{bold}, and the second-best is \underline{underlined}.}
\label{tab:all}
\end{table}

\begin{figure}[htbp]
    \centering

    \begin{subfigure}{0.48\textwidth}
        \centering
        \includegraphics[width=\linewidth]{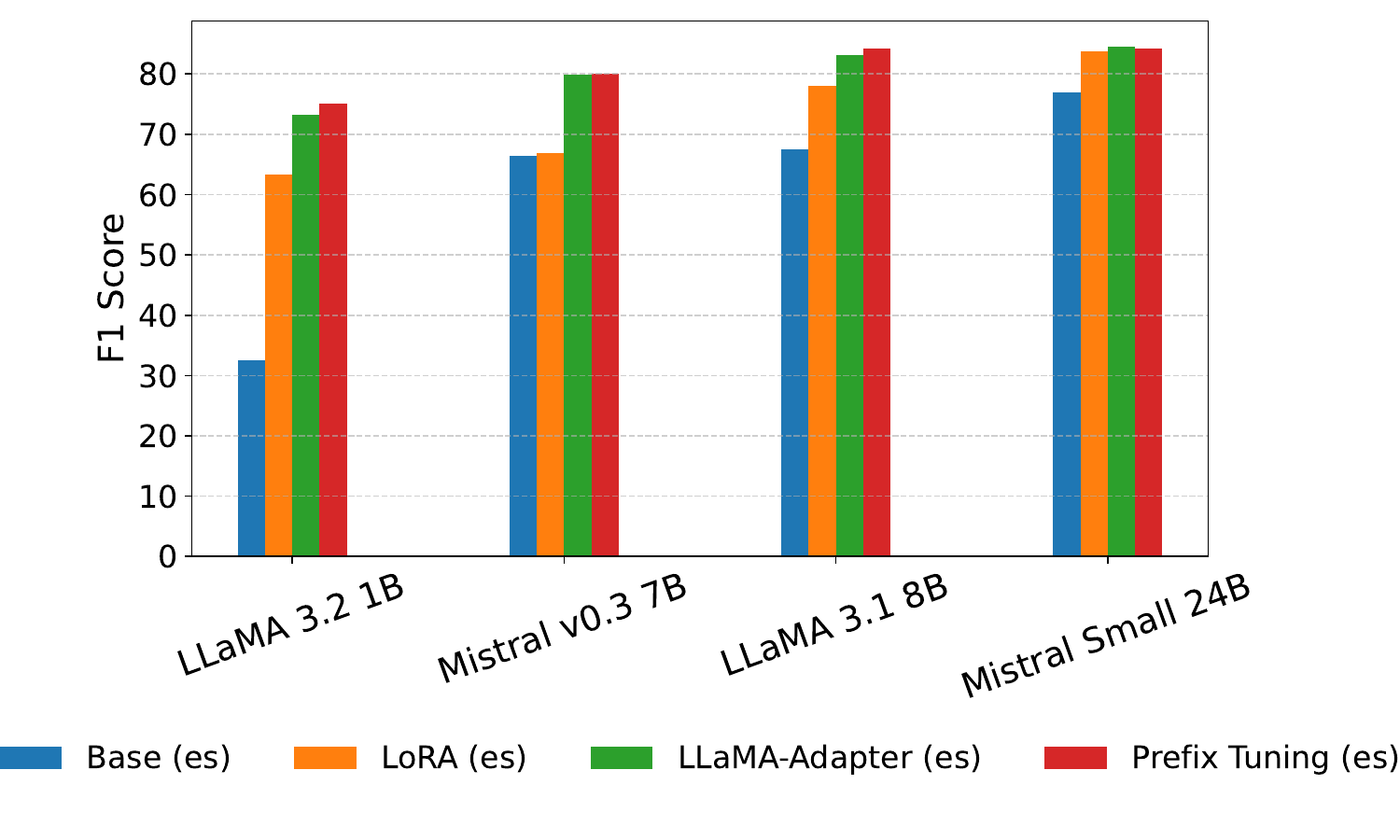}
        \caption{Comparison for Spanish.} 
        \label{fig:es}
    \end{subfigure}
    \hfill 
    \begin{subfigure}{0.48\textwidth}
        \centering
        \includegraphics[width=\linewidth]{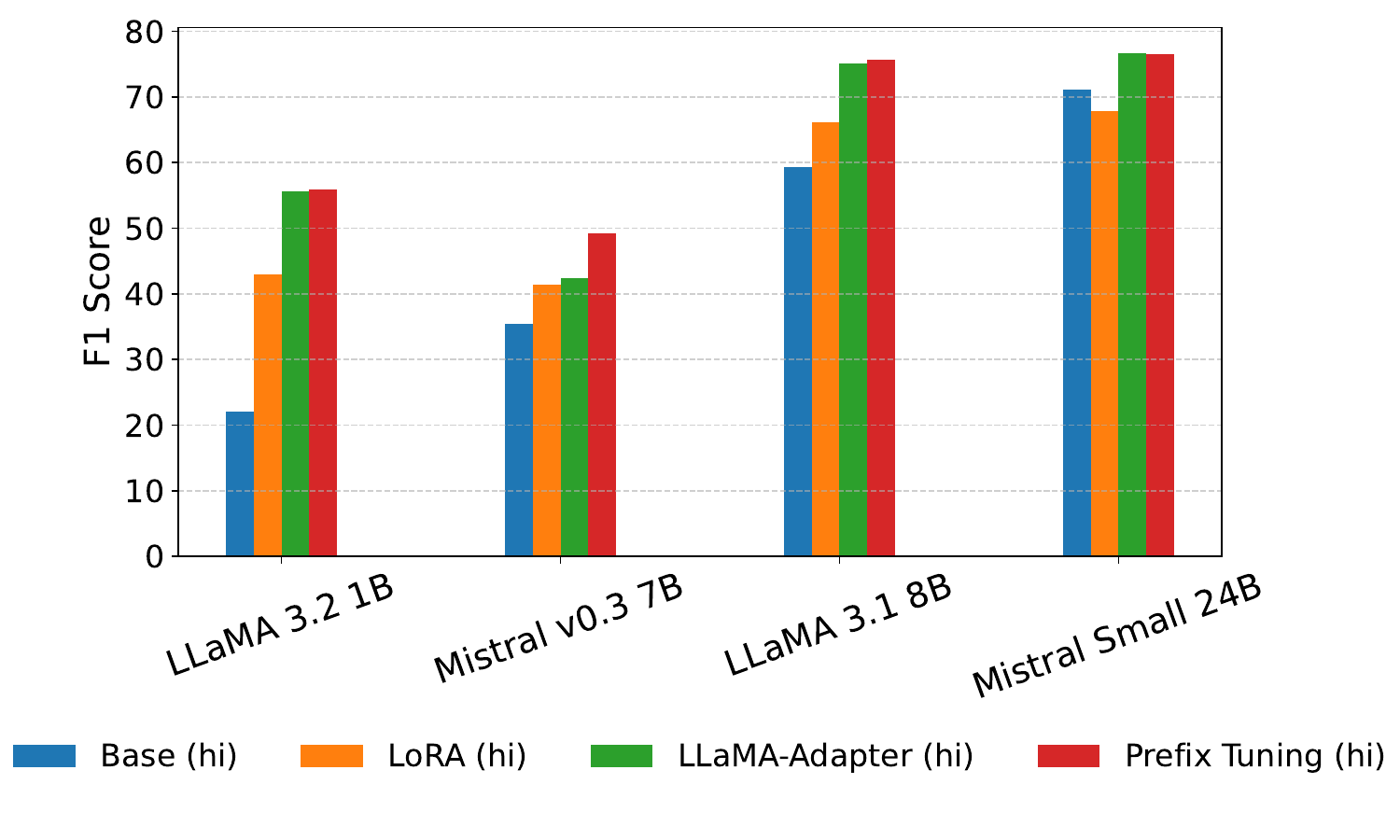}
        \caption{Comparison for Hindi.} 
        \label{fig:hi}
    \end{subfigure}

    \caption{Comparison of prefix-based methods across model sizes against LoRA fine-tuning on XQUAD (F1 score).}
    \label{fig:main_comparison}
\end{figure}

\paragraph{Effect of model size on prefix-based adaptation vs. LoRA.}
In Figures~\ref{fig:hi} and \ref{fig:es}, we compare the performance of prefix-based methods against LoRA on XQUAD for Spanish and Hindi across different model sizes. We observe that both prefix tuning and LLaMA Adapter consistently outperform LoRA across all model size variations in both languages. Table \ref{tab:all} shows that prefix-based adaptations scale more effectively with model size, maintaining their advantage even as the underlying model grows larger. In particular, prefix tuning yields consistent improvements, thus highlighting the robustness of prefix-based approaches for multilingual transfer.

\paragraph{Effectiveness of prefix-based methods on MGSM.}
Table~\ref{tab:mgsm} presents results on the MGSM benchmark with Llama-3.1 8B. LLaMA Adapter and prefix tuning consistently outperform LoRA, with prefix tuning achieving the best average score (+4\% over the base model). However, performance degraded for very low-resource languages like Swahili, Telugu, and Bengali. This suggests that while effective, prefix-tuning may not transfer well for complex reasoning and generation tasks without some language-specific data.



\begin{figure}
    \centering
    \begin{minipage}[t]{0.48\linewidth}
    \includegraphics[width=1\linewidth]{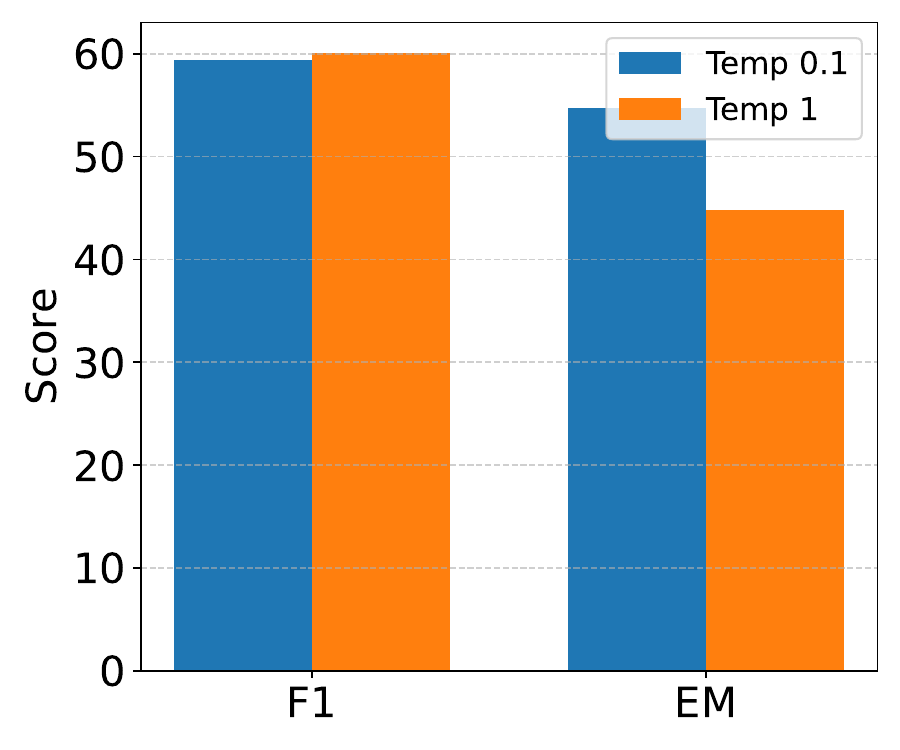}
    \label{fig:temperature}
    \end{minipage}
     \hfill
    \centering
    \begin{minipage}[t]{0.48\linewidth}
    \includegraphics[width=1\linewidth]{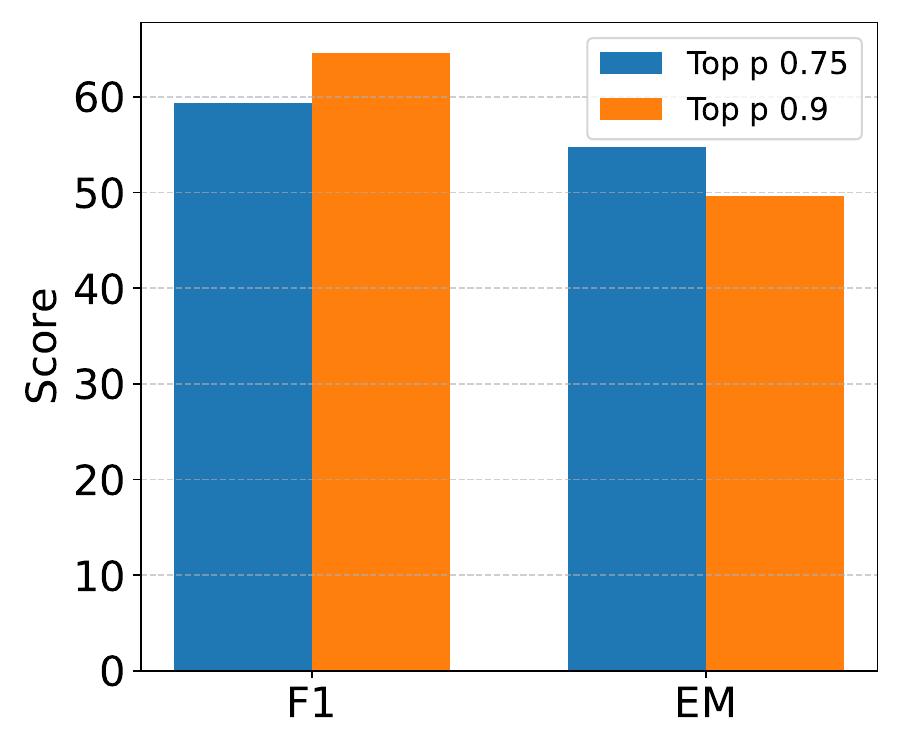}
    \end{minipage}
    \caption{Varying temperature (left) and top-p (right) values using Llama 3.2 (1B) on the XQUAD task.}
    \label{fig:topp}
\end{figure}

\paragraph{Varying temperature/top-p during prefix-tuning.}
For XQUAD, we have calculated both EM (Exact match) and F1 score. From figure \ref{fig:topp}, we find that while higher temperatures and top-p values can improve F1 scores on XQUAD, they often lead to a noticeable drop in EM. This highlights a trade-off between generating more diverse predictions (captured by F1) and producing exact matches (captured by EM). The best overall trade-off is obtained at our chosen setting of temperature=0.1 and top-p=0.75.



\captionsetup{font=small}
\begin{table}[t]
\centering
\scriptsize
\setlength{\tabcolsep}{4pt}
\begin{tabular}{c|c|c|c}
\toprule
\textbf{Method} & \textbf{Params} & \textbf{XNLI Acc.} & \textbf{XQUAD F1 Score} \\
\midrule
LoRA$_4$       & 2.36M & 74.4 & 74.1 \\
LoRA$_{128}$     & 75.50M & 76.7 & 76.0 \\
Llama Adapter     & 1.23M & 78.5 & \textbf{78.1}\\
Prefix tuning     & 1.23M & \textbf{78.7} & \textbf{78.1}\\
\bottomrule
\end{tabular}
\vspace{0.5em}
\caption{Higher Lora rank vs prefix based methods performance on XNLI and XQUAD for Llama 3.1 8B}
\label{tab:lora_rank_XNLI}
\end{table}

\paragraph{Performance comparison of LoRA$_4$, LoRA$_{r128}$ with prefix tuning and Llama Adapters.}  
\label{sec:lora_compare}
Table~\ref{tab:lora_rank_XNLI} provides a comparative analysis of LoRA fine-tuning under two rank configurations, $r=4$ and $r=128$, against prefix tuning and Llama adapters. While increasing the LoRA rank from 4 to 128 substantially increases the number of trainable parameters, the resulting performance improvements are relatively modest. More importantly, our results show that parameter-efficient prefix-based approaches namely prefix tuning and Llama adapters consistently outperform LoRA, even at higher ranks. This trend is evident in both the \textsc{XNLI} and \textsc{XQUAD} benchmarks, emphasizing the effectiveness of prefix-based adaptation for cross-lingual transfer. These findings suggest that simply scaling LoRA with larger ranks does not necessarily close the performance gap with prefix-based methods, and the latter remains a more efficient choice for multilingual scenarios.  

\paragraph{Impact of hyperparameter tuning on prefix-based adaptation.}  
\label{sec:hyperparamete}
Prefix-based approaches are governed by two critical hyperparameters: the prefix length and the number of transformer layers in which the prefixes are inserted. In soft prompt tuning, the adaptation is constrained to the input layer, whereas in prefix tuning, prefixes can be injected across multiple layers of the model. To better understand the effect of these design choices, we systematically varied both hyperparameters. Our experiments reveal that adapting 30 out of 32 layers with a prefix length of 10 tokens provides the strongest gains across benchmarks, as summarized in Tables~\ref{tab:insertion-layers} and \ref{tab:zero-init}. These results highlight the sensitivity of prefix-based methods to hyperparameter configurations, and emphasize the importance of carefully selecting the number of adapted layers and prefix length to maximize performance.(For results on other models, refer to Appendix~\ref{parameter}.)  

\section{Conclusion}
We show that prefix-based adaptation methods are a practical and efficient mechanism for cross-lingual transfer in decoder-only LLMs. Methods like soft prompting, prefix-tuning, and Llama adapters introduce learnable prefixes at different layers, while using relatively small numbers of trainable parameters. This leads to highly efficient, task-specific cross-lingual learning. 

Crucially, this performance was achieved using only English training data. We hypothesize this success stems from learning language-agnostic behaviors. By adding context vectors while keeping the base model frozen, these methods preserve the LLM's inherent multilingual capabilities. In contrast, methods that alter full model weights (e.g., full fine-tuning and LoRA) suffer from catastrophic forgetting when adapted monolingually, degrading performance in unseen languages. These findings advocate for prefix-based adaptation as a robust strategy for zero-shot cross-lingual transfer.

\begin{table}[t]
\centering
\begin{minipage}[t]{0.45\linewidth}
\centering
\small
\setlength{\tabcolsep}{5pt}
\begin{tabular}{c|c|c}
\toprule
\textbf{Layers} & \textbf{Params} & \textbf{Acc.} \\
\midrule
20 & 0.82M & 74.5 \\
30 & 1.23M & \textbf{78.7} \\
32 & 1.31M & 78.0 \\
\bottomrule
\end{tabular}
\vspace{0.5em}
\caption{\textsc{XNLI} performance accuracy by varying number of Llama 3.1 8B layers in which prefixes are inserted.}
\label{tab:insertion-layers}
\end{minipage}
\hfill
\begin{minipage}[t]{0.45\linewidth}
\centering
\small
\setlength{\tabcolsep}{4pt}
\begin{tabular}{c|c|c}
\toprule
\textbf{Tokens} & \textbf{Params} & \textbf{Acc.} \\
\midrule
5 & 0.61M & 77.8 \\
10 & 1.23M & \textbf{78.7} \\
20 & 2.46M & 76.0 \\
\bottomrule
\end{tabular}
\vspace{0.5em}
\caption{\textsc{XNLI} performance accuracy by varying number of prefix tokens in 30 Llama 3.1 8B layers.}
\label{tab:zero-init}
\end{minipage}
\end{table}

\section*{Limitations}
Our study shows that prefix-based methods yield strong zero-shot cross-lingual performance, but it has several limitations. First, due to computational constraints, our experiments were limited to 24B models; extending to larger models is a promising direction for future work. Second, our evaluations used only English as the source language. Analyzing other source languages could offer deeper insights into the methods' cross-lingual capabilities. Finally, due to computational constraints, we were unable to perform an extensive hyperparameter search for full fine-tuning. We would like to emphasize this limitation more explicitly and clarify that our intention is not to claim full fine-tuning is inherently weaker, but rather to highlight that parameter-efficient methods provide strong alternatives under realistic computational constraints. In future work, we plan to explore improving response generation for low-resource languages as seen in the MGSM benchmark and also explore more diverse response generation tasks (e.g. summarization and translation). We also plan to investigate why prefix-tuning is effective through attention visualization and representation probing.

\section*{Acknowledgments}
We are grateful to the anonymous reviewers
for their insightful feedback. The last author gratefully acknowledges the generous support provided by the joint AI/ML initiative of Amazon and the Indian Institute of Technology Bombay.

\bibliography{custom}

\appendix

\section{Prompt Templates}
\label{sec:prompt_template}
Training and inference prompts for all the three benchmarks we have evaluated. For MGSM, we use the 8-shot chain-of-thought prompt as in \cite{wei2022chain} (maj@1) to evaluate. 

\begin{tcolorbox}[colback=white!5!white, colframe=black!75!black, sharp corners=southwest, boxrule=0.8pt, fonttitle=\bfseries, title=XQUAD]
Below is an instruction that describes a task, paired with an input that provides further context. Write a response that appropriately completes the request.

\textbf{\#\#\# Instruction:}

You will answer reading comprehension questions using information from a provided passage. Extract the exact answer from the passage without modification and present it in the following structured format:

\{'answer' : \texttt{<Extracted Answer>\}}

\textbf{\#\#\# Input:} \\
Context: \\
\texttt{<context>} \\
Question: \\
\texttt{<question>} \\

\textbf{\#\#\# Response:} \\
\texttt{\{'answer':}
\end{tcolorbox}

\begin{tcolorbox}[colback=white!5!white, colframe=black!75!black, sharp corners=southwest, boxrule=0.8pt, fonttitle=\bfseries, title=Belebele]
Below is an instruction that describes a task, paired with an input that provides further context. Write a response that appropriately completes the request.

\textbf{\#\#\# Instruction:}

The task is to perform a reading comprehension task. Given the following passage, question, and answer choices, output the number corresponding to the correct answer only.

\textbf{\#\#\# Input:} \\
Passage: \\
\texttt{<passage>} \\
Question: \\
\texttt{<question>} \\
Choices: \\
\texttt{<choices>} \\

\textbf{\#\#\# Response:} The correct choice number is
\end{tcolorbox}

\begin{tcolorbox}[colback=white!5!white, colframe=black!75!black, sharp corners=southwest, boxrule=0.8pt, fonttitle=\bfseries, title=\textsc{XNLI}]
Below is an instruction that describes a task, paired with an input that provides further context. Write a response that appropriately completes the request.

\textbf{\#\#\# Instruction:}

The task is to solve Natural Language Inference (NLI) problems. NLI is the task of determining whether the inference relation between the second sentence (Hypothesis) with respect to the first sentence (Premise) is one of the following: \\
1. Entailment \\
2. Neutral \\
3. Contradiction \\
Output the relation number only.

\textbf{\#\#\# Input:} \\
Premise: \\
\texttt{<premise>} \\
Hypothesis: \\
\texttt{<hypothesis>} \\

\textbf{\#\#\# Response:} The relation number is
\end{tcolorbox}

\section{Languages details}
Evaluation language details included in the benchmarks are given in Tables \ref{tab:xnli_iso_codes}, \ref{tab:lang_family} and \ref{tab:lang_script}. 
\label{sec:language_detail}

\begin{table}[t]
\centering
\begin{tabular}{c|c}
\hline
\textbf{Benchmark} & \textbf{Languages} \\
\hline
XNLI  & en, hi, el, vi, sw, bg, th, ar,  \\
& ar, de, es, fr, ru, tr, zh, ur \\
\hline
XQUAD & en, hi, el, vi, ar, de, es, ro,  \\
& ru, th, tr, zh\\
\hline
\end{tabular}
\caption{Languages used in the XNLI and XQUAD benchmarks.}
\label{tab:xnli_iso_codes}
\end{table}

\section{Hyperparameter details}
\label{parameter}
We insert 10 prefix tokens across 30 layers for LLaMA 3.1 8B, Mistral 7B, and Mistral 24B, while for LLaMA 3.2 1B, the tokens are inserted across all layers as it is small.For full fine-tuning, we used a batch size of 8, a learning rate of 1e-5 with a cosine learning rate scheduler, a warm-up ratio of 0.1, and trained the model for 2 epochs.Finally for LoRA fine tuning, we applied it to the Q, K, and V projection matrices across all layers.

\begin{table}[h]
\centering
\begin{tabular}{lcc}
\hline
\textbf{Language} & \textbf{F1} & \textbf{EM} \\
\hline
ar  & 14.43 &  9.83 \\
de  & 61.95 & 43.36 \\
el  & 22.63 & 17.98 \\
en  & 84.69 & 72.10 \\
es  & 62.08 & 41.34 \\
hi  & 15.23 & 11.76 \\
ro  & 58.57 & 40.76 \\
ru  & 18.65 & 10.42 \\
th  & 16.13 & 12.35 \\
tr  & 42.38 & 26.72 \\
vi  & 43.47 & 25.88 \\
zh  & 12.66 &  9.50 \\
\hline
\textbf{Avg} & \textbf{37.74} & \textbf{26.83} \\
\hline
\end{tabular}
\caption{Full fine tuning performance of Llama 3.1 8B on XQUAD}
\label{tab:fft_3}
\end{table}

\begin{table}[t]
\centering
\small
\begin{tabular}{l|l}
\toprule
\textbf{Language} & \textbf{Family} \\
\midrule
Kazakh             & Turkic \\
Kyrgyz             & Turkic \\
North Azerbaijani  & Turkic \\
Kannada            & Dravidian \\
Malayalam          & Dravidian \\
Tamil              & Dravidian \\
Amharic            & Afro-Asiatic \\
Tigrinya           & Afro-Asiatic \\
Tsonga             & Afro-Asiatic \\
Sindhi             & Indo-Aryan \\
Odia               & Indo-Aryan \\
Sinhala            & Indo-Aryan \\
Russian            & Balto-Slavic \\
Serbian            & Balto-Slavic \\
Slovak             & Balto-Slavic \\
\bottomrule
\end{tabular}
\caption{Languages grouped by family included in Belebele}
\label{tab:lang_family}
\end{table}

\begin{table}[t]
\centering
\small
\begin{tabular}{ll}
\toprule
\textbf{Language} & \textbf{Script} \\
\midrule
Kyrgyz             & Cyrillic \\
Russian            & Cyrillic \\
Serbian            & Cyrillic \\
Burmese            & Burmese \\
Shan               & Burmese \\
Swati              & Latin \\
Sundanese          & Latin \\
Bambara            & Latin \\
Sindhi             & Arabic \\
Egyptian Arabic    & Arabic \\
Western Persian    & Arabic \\
Amharic            & Ethiopic \\
Tigrinya           & Ethiopic \\
\bottomrule
\end{tabular}
\caption{Languages grouped by script included in Belebele}
\label{tab:lang_script}
\end{table}

\section{Complete elaborated experiment results
}
\label{sec:Final_numbers}

\begin{table*}[t]
\centering
\small
\resizebox{\textwidth}{!}{
\begin{tabular}{l|ccccccccccccccc|c}
\toprule
\textbf{Method} & \textbf{en} & \textbf{hi} & \textbf{el} & \textbf{vi} & \textbf{sw} & \textbf{bg} & \textbf{th} & \textbf{ar} & \textbf{de} & \textbf{es} & \textbf{fr} & \textbf{ru} & \textbf{tr} & \textbf{zh} & \textbf{ur} & \textbf{Avg} \\
\midrule
Base Model & 34.3 & 34.7 & 33.8 & 33.6 & 33.4 & 33.8 & 32.8 & 33.6 & 34.1 & 33.8 & 33.5 & 33.6 & 34.2 & 33.9 & 33.6 & 33.8\\
LoRA$_4$ & 47.3 & 42.2 & 42.1 & 44.8 & 43.5 & 47.3 & 44.0 & 45.0 & 41.6 & 42.1 & 40.0 & 48.3 & 40.0 & 43.9 & 40.6 & 43.5\\
Soft Prompts & 79.4 & 41.8 & 46.7 & 67.6 & 44 & 70.5 & 48.8 & 56.5 & 73.2 & 75.3 & 75.9 & 67.0 & 60.9 & 69.4 & 49.5 & 61.7\\  
Llama Adapter & \textbf{92.0} & \textbf{58.1} & \textbf{64.8} & \textbf{69.3} & \textbf{46.9} & \underline{73.9} & \underline{61.3} & \textbf{61.7} & \textbf{79.0} & \underline{79.3} & \textbf{80.6} & \underline{76.0} & \underline{65.2} & \underline{76.7} & \textbf{55.6} & \textbf{69.4}\\

Prefix Tuning & \underline{90.8} & \underline{56.7} & \underline{61.9} & \textbf{69.3} & \underline{43.4} & \textbf{75.7} & \textbf{62.8} & \underline{61.5} & \underline{78.8} & \textbf{80.3} & \underline{79.5} & \textbf{76.7} & \textbf{63.9} & \textbf{78.3} & \underline{54.5} & \underline{69.0} \\
\bottomrule
\end{tabular}
}
\caption{Mistral v0.3 7B performance (accuracy) on XNLI benchmark comparing LoRA and prefix based adaption methods.The best performance for each language is shown in \textbf{bold}, and the second-best is \underline{underlined}.}
\label{tab:task-a}
\end{table*}



\begin{table*}[t]
\centering
\small
\begin{tabular}{l|cccccccccccc|c}
\toprule
\textbf{Method} & \textbf{en} & \textbf{hi} & \textbf{el} & \textbf{vi} & \textbf{ar} & \textbf{de} & \textbf{es} & \textbf{ro} & \textbf{ru} & \textbf{th} & \textbf{tr} & \textbf{zh} & \textbf{Avg} \\
\midrule
Base Model & 77.7 & 35.4 & 47.9 & 62.7 & 46.9 & 65.4 & 66.4 & 64.3 & 53.8 & \underline{47.4} & 48.0 & 57.8 & 56.1\\
LoRA$_4$ & 82.5 & 41.37 & \underline{53.52} & 48.0 & \underline{54.1} & 67.1 & 68.2 & 66.7 & 58.8 & \textbf{53.2} & 51.1 & 64.9 & 59.12\\
Soft Prompts & 72.1 & 1.6 & 19.4 & 42.2 & 18.4 & 61.6 & 62.3 & 59.6 & 49.3 & 10.1 & 48.1 & 18.4 & 38.6\\
Llama Adapter & \textbf{88.5} & \underline{42.5} & 53.4 & \underline{69.1} & 51.1 & \underline{75.9} & \underline{80.0} & \textbf{78.6} & \textbf{72.3} & 41.3 & \underline{58.3} & \textbf{71.0} & \underline{65.1}\\
Prefix Tuning & \underline{88.4} & \textbf{49.3} & \textbf{60.4} & \textbf{69.5} & \textbf{55.4} & \textbf{77.4} & \textbf{80.0} & \underline{78.2} & \underline{71.7} & 46.1 & \textbf{60.9} & \underline{69.1} & \textbf{67.2} \\
\bottomrule
\end{tabular}
\caption{Mistral v0.3 7B performance (F1 score) on XQUAD benchmark comparing LoRA and prefix based adaption methods.The best performance for each language is shown in \textbf{bold}, and the second-best is \underline{underlined}.}
\label{tab:task-c1}
\end{table*}

\begin{table*}[t]
\centering
\begin{minipage}{0.49\textwidth}
\centering
\resizebox{\textwidth}{!}{
\begin{tabular}{l|l|ccccc}
\toprule
\textbf{Script} & \textbf{Language} & \textbf{Base} & \textbf{LoRA$_4$} & \textbf{Soft} & \textbf{Llama} & \textbf{Prefix} \\
\textbf{} & \textbf{} & \textbf{Model} & \textbf{} & \textbf{Prompt} & \textbf{Adapter} & \textbf{tuning} \\
\midrule
\multirow{3}{*}{Cyrillic} 
 & Kyrgyz    & 31.7 & 29.2 & 35.8 & \underline{34.1} & \textbf{35.5} \\
 & Russian   & 57.3 & 62.2 & \underline{83.1} & \textbf{83.8} & 82.3 \\
 & Serbian   & 55.5 & 60.2 & \underline{79.0} & \textbf{79.8} & 76.5 \\
\midrule
\multirow{3}{*}{Burmese} 
 & Burmese    & 28.3 & 23.0 & \textbf{33.0} & \underline{30.8} & 30.7 \\
 & Shan   & 26.0 & 21.5 & \underline{26.1} & 25.3 & \textbf{27.0} \\
\midrule
\multirow{3}{*}{Latin} 
 & Swati    & 28.6 & 27.3 & 29.6 & \underline{30.0} & \textbf{32.0} \\
 & Sundanese   & 32.1 & 30.5 & \textbf{37.4} & \underline{35.7} & 35.4 \\
 & Bambara   & 29.3 & 28.3 & \underline{31.3} & 31.2 & \textbf{32.8} \\
\midrule
\multirow{3}{*}{Arabic} 
 & Sindhi    & 31.3 & 24.3 & \textbf{31.4} & 29.2 & 30.8 \\
 & Egyptian Arabic   & 39.3 & 35.0 & \textbf{48.6} & \underline{45.1} & 43.7 \\
 & Western Persian   & 41.2 & 35.1 & \textbf{55.4} & 49.8 & \underline{52.5} \\
\midrule
\multirow{3}{*}{Ethiopic}
 & Amharic     & 29.3 & 22.7 & \textbf{31.1} & 29.2 & \underline{30.7} \\
 & Tigrinya    & 28.3 & 23.0 & 25.7 & 26.1 & 27.0 \\
\bottomrule
\end{tabular}
}
\caption{Performance (accuracy) of Mistral v0.3 7B on the Belebele benchmark, grouped by language \textbf{script}, comparing LoRA and prefix-based adaptation methods.The best performance for each language is shown in \textbf{bold}, and the second-best is \underline{underlined}.}
\label{tab:family_mistral}
\end{minipage}
\hfill
\begin{minipage}{0.49\textwidth}
\centering
\resizebox{\textwidth}{!}{
\begin{tabular}{l|l|ccccc}
\toprule
\textbf{Family} & \textbf{Language} & \textbf{Base} & \textbf{LoRA$_4$} & \textbf{Soft} & \textbf{Llama} & \textbf{Prefix} \\
\textbf{} & \textbf{} & \textbf{Model} & \textbf{} & \textbf{Prompting} & \textbf{Adapter} & \textbf{tuning} \\
\midrule
\multirow{3}{*}{Turkic}
 & Kazakh            & 33.7 & 29.4 & \textbf{38.0} & 34.3 & \underline{35.6} \\
 & Kyrgyz             & 31.7 & 29.2 & \textbf{35.8} & 34.1 & \underline{35.5} \\
 & North Azerbaijani & 34.7 & 35.2 & \textbf{45.5 }& \underline{42.3} & 42 \\
\midrule
\multirow{3}{*}{Dravidian}
 & Kannada           & 34.3 & 25.7 & \textbf{38.1} & 34.2 & \underline{36} \\
 & Malayalam         & 31.8 & 25.7 & \textbf{36.7} & \underline{31.8} & 31.4 \\
 & Tamil             & 34.1 & 29.0 & \underline{39.8} & 36.5 & \textbf{40.0} \\
\midrule
\multirow{3}{*}{Afro-Asiatic}
 & Amharic            & 29.3 & 22.7 & \textbf{31.1} & 29.2 & \underline{30.7} \\
 & Tigrinya            & 28.3 & 23.0 & 25.7 & 26.1 & 27.0 \\
 & Tsonga      & 28.4 & 28.5 & \textbf{34.7}  & 33.3 & \underline{33.8} \\
\midrule
\multirow{3}{*}{Indo-Aryan}
 & Sindhi            & 31.3 & 24.3 & \textbf{31.4} & 29.2 &30.8 \\
 & Odia            & 30.5& 23.6 & 30.3 & \textbf{30.7} & \underline{30.5} \\
 & Sinhala & 32.2 & 27.1 & 34.5 & 29.4 & 34.5 \\
 \midrule
\multirow{3}{*}{Balto-Slavic}
 & Russian   & 57.3 & 62.2 & \underline{83.1} & \textbf{83.8} & 82.3 \\
 & Serbian   & 55.5 & 60.2 & \underline{79.0} & \textbf{79.8} & 76.5 \\
 & Slovak    & 52.9 & 58.2 & \textbf{73.1} & \underline{72.8} & 72.3 \\
\bottomrule
\end{tabular}
}
\caption{Performance (accuracy) of Mistral v0.3 7B on the Belebele benchmark, grouped by language \textbf{family}, comparing LoRA and prefix-based adaptation methods.The best performance for each language is shown in \textbf{bold}, and the second-best is \underline{underlined}.}
\label{tab:script_mistral}
\end{minipage}
\end{table*}

\end{document}